\documentclass[a4paper,twoside]{article}
\pdfoutput=1
\usepackage{epsfig}
\usepackage{subfigure}
\usepackage{calc}
\usepackage{amssymb}
\usepackage{amstext}
\usepackage{amsmath}
\usepackage{amsthm}
\usepackage{multicol}
\usepackage{pslatex}
\usepackage{apalike}
\usepackage{graphicx}
\usepackage{balance} 
\usepackage[english]{babel}
\usepackage{blindtext}
\usepackage{float}
\usepackage{epstopdf}
\usepackage{balance}
\usepackage{url}

\usepackage{SCITEPRESS}     

\newcommand{\kmafShortName}{KMARF}
\newcommand{\kmafLongName}{Knowledge Management and Automated Reasoning Framework}

\begin{document}

\title{A Framework for Knowledge Management and Automated Reasoning Applied on Intelligent Transport Systems}

\author{\authorname{Aneta Vulgarakis Feljan\sup{1}, Athanasios Karapantelakis\sup{1}, Leonid Mokrushin\sup{1}, Hongxin Liang\sup{1}, Rafia Inam\sup{1}, Elena Fersman\sup{1}, Carlos R.B. Azevedo\sup{2}, Klaus Raizer\sup{2} and Ricardo S. Souza\sup{2}}
\affiliation{\sup{1}Ericsson Research, Sweden}
\affiliation{\sup{2}Ericsson Research - ARBB, Brazil}
\email{Email: fistname.lastname@ericsson.com}
}

\keywords{Knowledge model, Reasoning, Ontology, Model transformation, Intelligent Transport Systems}

\abstract{Cyber-Physical Systems in general, and Intelligent Transport Systems (ITS) in particular use heterogeneous data sources combined with problem solving expertise in order to make critical decisions that may lead to some form of actions e.g., driver notifications, change of traffic light signals and braking to prevent an accident. Currently, a major part of the decision process is done by human domain experts, which is time-consuming, tedious and error-prone. Additionally, due to the intrinsic nature of knowledge possession this decision process cannot be easily replicated or reused. Therefore, there is a need for automating the reasoning processes by providing computational systems a formal representation of the domain knowledge and a set of methods to process that knowledge. In this paper, we propose a knowledge model that can be used to express both declarative knowledge about the systems' components, their relations and their current state, as well as procedural knowledge representing possible system behavior. In addition, we introduce a framework for knowledge management and automated reasoning (KMARF). The idea behind KMARF is to automatically select an appropriate problem solver based on formalized reasoning expertise in the knowledge base, and convert a problem definition to the corresponding format. This approach automates reasoning, thus reducing operational costs, and enables reusability of knowledge and methods across different domains. We illustrate the approach on a transportation planning use case. }
	

\onecolumn \maketitle \normalsize \vfill



\section{\uppercase{Introduction}}
\label{introduction}



\noindent 
Intelligent Transport Systems (ITS) present a subclass of Cyber-Physical Systems (CPS) due to the interaction between physical systems (vehicles) and distributed information acquisition and propagation infrastructure (wired/wireless networks, sensors, actuators, processors and software). They are designed to offer innovative services to transportation network users and managers alike, and cover a broad area of advanced applications, ranging from improving management and safety of the transportation network, to infotainment services~\cite{karagiannis2011vehicular}. 
ITS in general use a vast amount of heterogeneous data streams and information (e.g., behavioral models) that need to be analyzed, combined and actioned upon, which creates a complexity that goes far beyond mere human management. Hence, automation is a must, which in its turn requires formalization of knowledge extracted from the different sources such as sensor networks, documents, tools and from system experts. Moreover, in order to be able to provide support for information interoperability between applications, ITS-related information needs to be semantically enriched in other words identified, categorized and described explicitly. Even though standardization activities have facilitated certain network-level interoperability~\cite{ieee11802} and Quality of Service (QoS), there are still inconsistencies between ITS groups on how data are modeled on a semantic level, as each organization (e.g., government regulator, commercial provider, academic institution) has created domain-specific information models for its respective ITS applications. 

In order to build intelligent systems one needs to start by modeling and formalizing the knowledge that exists (such as domain concepts, behaviors and context) and is retrieved from human experts. Knowledge representation and reasoning (KR\&R) is a field in artificial intelligence that aims at building intelligent systems that know about their world and are able to automatically draw conclusions and act upon them, as humans do~\cite{baral2003knowledge}.  A fundamental assumption in KR\&R is that knowledge is represented in a tangible form (usually via ontologies), suitable for processing by dedicated reasoning engines. However, up to today, there is a lack of frameworks for general intelligence that will solve several classes of reasoning problems, such as planning, verification and optimization.  In addition, it is still not clear to the intelligent software community how to effectively cope with the integration of both declarative and procedural knowledge and some authors advocate for keeping the behavior descriptions separated from the semantic, static domain knowledge~\cite{ghallab_automated_planning_2014}.

In this paper, we present a \kmafLongName{} (\kmafShortName{}), which  targets multiple reasoning problems. The purpose for conceiving \kmafShortName{} is to (a) reduce  system development and deployment time, by reusing as much knowledge as possible, such as domain models, behaviors and reasoning mechanisms, and 
(b) reduce operational costs by requiring less human involvement in system operations. 
The strong point of \kmafShortName{} is that it relies on a knowledge model that combines both declarative and procedural knowledge. This combination allows us to do extensive analysis and provide answers to different reasoning questions, such as ``what is the optimal strategy for reaching a desired system state?'' or ``which actions have led the system to a given state?''. In order to use problem solvers specialized in solving specific classes of reasoning problems, \kmafShortName{} can be extended with model transformation rules that translate from our knowledge model to the targeted problem solver model.

\kmafShortName{} is specifically targeting CPS, but it can be applied in other areas as well. In this paper, we focus on one of the CPS domains where \kmafShortName{} may be applied -- the ITS domain. As an illustration example, we consider a transportation planning problem i.e.,  transport passengers or goods with a minimal cost. A cost can be e.g., the traveled distance, the time needed for transportation of each of the passengers or goods, or the number of buses or trucks required for transportation. The answer to the task may be a plan i.e., a sequence of steps for the system to perform in order to reach the goal state. In case the task cannot be performed the answer from \kmafShortName{} could be a reason why the task cannot be performed, as well as the possible solution, which could be to increase the number of vehicles.

In brief, our contribution is fourfold:
\begin{itemize}
	
	\item{A model for representing both declarative and procedural knowledge from CPS and ITS domains in a machine-readable form (Section~\ref{knowledgeModel})}.
	\item{An architecture of a generic framework for knowledge management and automated reasoning (\kmafShortName{}) (Section~\ref{architecture}), relying on the introduced knowledge model.}
	\item{A taxonomy of ITS domain concepts that can be used when reasoning about ITS specific problems (Section~\ref{domainModeling}).}
	\item{A prototype implementation of the \kmafShortName{} architecture (Section~\ref{implementation}) and an evaluation of the benefits of \kmafShortName{} in terms of reducing the effort required for formalizing the knowledge and performing the reasoning based on it (Section~\ref{evaluation}).}

	
\end{itemize}




\section{\uppercase{\kmafShortName{} -- Framework for Knowledge Management and Automated Reasoning}}
\label{KMAF}


\noindent In this section, we introduce our \kmafLongName{} (\kmafShortName{}) by showing its architecture and the knowledge model that it relies on. \kmafShortName{} is targeting multiple reasoning problem classes (such as  planning, verification and optimization) that can share the same underlying state representation. This enables reusability of knowledge and methods across different domains.

\subsection{The Knowledge Model}
\label{knowledgeModel}
\paragraph{Syntax}

We model declarative knowledge by describing discrete states of a system.
One such state may represent the current state, and the others may describe
either previous system states or hypothetical states that the system may end
up in the future. In this context we do not strictly apply the notion of time,
i.e., the system may change its state instantly. However, the order of states
is important as it describes how the system evolves and may explain the reasons
behind its progress.

\begin{figure}
    \centering
    \includegraphics[width=76mm]{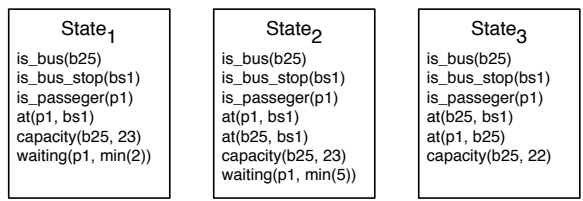}
    \caption{Specification of three example states\label{fig-states}.}
\end{figure}

A \textit{state} is represented by an (implicitly conjunctive) set of predicates
$\{P_1, P_2, ...\}$ expressing the facts known about the state. Each predicate
is a compound term that has a form of $P_i(a_1, a_2, ..., a_n)$, where $P_i$
is a predicate's functor specified as a literal, i.e., a sequence of characters,
and $a_j | j\in[1..n]$ are the arguments. The number of arguments $n$ is called arity
of the predicate. If $n=0$, then $P_i$ denotes a simple atomic fact. If $n>0$,
then $P_i$ denotes a factual relation between its $n$ arguments. The arguments
of predicates may include:

\begin{itemize}
    \item numbers denoting literal quantity;
    \item literals denoted by sequence of alphanumeric characters that start
          from a lower case character $\{a,b,c,...\}$ that represent objects
          or concepts in the domain (e.g., $car$ may represent ``a car'' );
    \item compound terms of the form $T_i(a_1, a_2, ..., a_n)$ that may have
          one or many arguments, which can be numbers, literals or compound
          terms (e.g., $velocity(car, kmh(50))$ may stand for ``the velocity
          of the car $car$ is 50 km/h'').
\end{itemize}

An example shown in Figure~\ref{fig-states} contains a state $State_1$ defined
using a set of six predicates $\{is\_bus, is\_bus\_stop, is\_passeger, at,
capacity, waiting\}$. The arguments of the first three predicates declare
existence of bus $b25$, bus stop $bs1$, and passenger $p1$ accordingly. The
fourth predicate states that passenger $p1$ is at bus stop $bs1$. The fifth
predicate indicates that bus $b25$ has 23 available places. The last predicate
declares a fact that passenger $p1$ has been waiting at the bus stop from
last two minutes.

The procedural knowledge is modeled as a collection of specifications of potential
transitions between states. A \textit{transition} specification consists of a
\textit{precondition}, a \textit{computation}, and an \textit{action}. Precondition
and action both have the same syntax as the state, i.e., they are represented as a
set of predicates, except the following two differences. Variables denoted by
sequences of alphanumeric characters starting from a capital letter $\{X,Y,Z,...\}$
are allowed in arguments of predicates and compound terms in both precondition
and action. The action predicates are restricted to the set $\{add, delete\}$.
Intuitively, action predicates denote the procedures performed with a state when
a transition is performed. Computation is an ordered list of effect free function
calls, i.e. they do not modify the state and are only used during the processing of
the transition. The arguments of a function call may be numbers, literals, variables,
and functions. If a function returns a value, the last argument of a function call
is a variable that holds it. The result of a function call may be used as an
argument in subsequent function calls of the computation or in the action.

\begin{figure}
    \centering
    \includegraphics[width=76mm]{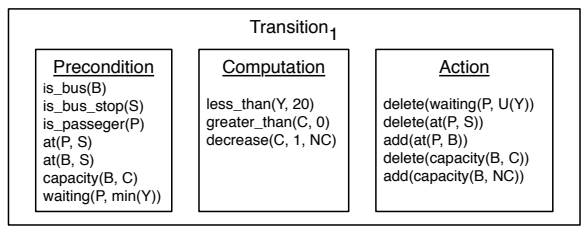}
    \caption{An example of a transition specification\label{fig-rules}.}
\end{figure}

Example in Figure~\ref{fig-rules} demonstrates specification of a transition
$Transition_1$ that allows a system to evolve from state $State_2$ to state
$State_3$ defined in Figure~\ref{fig-states}. After matching the precondition, the
computation checks if a passenger has been waiting for less than 20 minutes and if
there is enough capacity to onboard a passenger, it decreases the bus capacity
by 1. The action removes the passenger waiting predicate, updates passenger
location and available bus capacity value.


\begin{figure*}[!t]
	\centering	\includegraphics[width=0.96\textwidth]{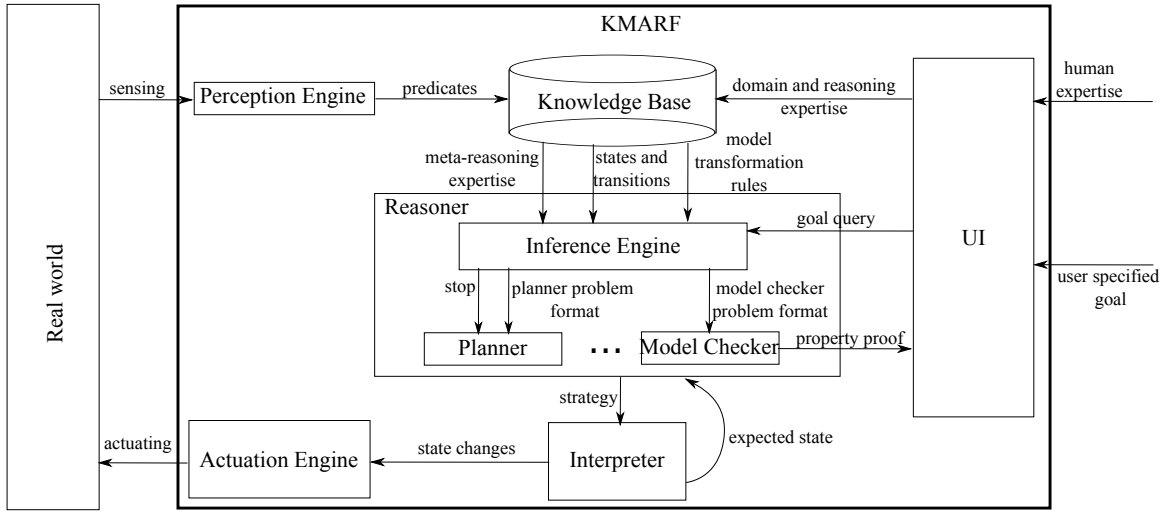}
	\caption{A high level, conceptual view of the architecture of \kmafShortName{}.\label{fig-architecture}}
\end{figure*}

\paragraph{Semantics} 
We formalize semantics of literals, compound terms and predicates by associating meanings to their symbols. For example, we use the predicate $at(p1, bs1)$ to model the fact that a passenger $p1$ is at the bus stop $bs1$. We say that the meaning of $at$ is to represent a close spacial relationship between its two arguments.  

We define semantics of our knowledge model in terms of a transition system. Formally, a transition system is a tuple $(S, T, \rightarrow)$, where $S$ is a set of states, $T$ is a set of transition specification names, and $\rightarrow$ is set of state transitions (i.e., a subset of $S \times T \times S$).  The fact that $(State_1, Transition_1, State_2) \in{ \rightarrow}$ is written as  $State_1 \xrightarrow{Transition_1} State_2$, and represents a transition between a source state $State_1$ and a destination state $State_2$ by applying transition specification $Transition_1$.

In order for a transition specification to be applied its precondition must match the source state and its computation must succeed. The semantics of matching the precondition with a state are formalized by defining a logical unification between predicates of the precondition and the predicates of the state, as follows. We unify every predicate in the precondition with all the predicates from the state, and use variables substitutions in subsequent unification of the remaining precondition predicates. If the unification of all precondition predicates with the state predicates succeeds, the computed variable substitution is used in the computation and the action, as explained below. Obviously, there can be multiple matches of a precondition with a source state. In this case, every match will produce a transition in $\rightarrow$ given that the corresponding computation succeeds.

The meaning of the computation is evaluation of its functions in the order of specification. We use denotational semantics to define the meaning of a function $F$ as a set of ordered tuples $$\{<a_1^1, ..., a_{n-1}^1,a_n^1>, ..., <a_1^m, ..., a_{n-1}^m,a_n^m>\},$$ where $a_1^i,...,a_{n-1}^i$ are function arguments, and $a_n^i$ is the value returned by $F$ given those arguments. A function call $F(a_1,...,a_{n-1},a_n)$ is the process of finding such $a_n$ for given $a_1,...,a_{n-1}$ that there is a tuple $<a_1^j, ..., a_{n-1}^j,a_n^j>$ in the definition of $F$ for some $j$. If there is no such tuple found, the function call fails. Otherwise, the function call succeeds and the value of $a_n^j$ is assigned to a corresponding variable. If the returned value $a_n$ is of boolean type, i.e. it belongs to the set $\{true, false\}$, then the function succeeds if $a_n = true$ and fails if $a_n = false$. A computation succeeds if all the function calls in it succeed. Otherwise, the computation fails.

The semantics of the transition action execution are defined by two operations. The first operation instantiates predicates in the action by applying computed variable substitution to them. This means that all the variables in the action predicates are replaced with the corresponding values from the variable substitution. The second operation copies all the predicates from the source state to the destination state, and for every predicate in the action we perform the following procedures on the destination state:

\begin{itemize}
	\item if the instantiated predicate symbol is $add$, then its argument is treated as a predicate, and it is added to the destination state;
	\item if the instantiated predicate symbol is $delete$, then its argument is treated as a predicate, and it is removed from the destination state.
\end{itemize}


\subsection{Architecture}
\label{architecture}
After introducing the knowledge model, we can move on describing \kmafShortName{}. A high-level conceptual view of the architecture of the framework is depicted in Figure~\ref{fig-architecture}. The main components of \kmafShortName{} are the \emph{Perception Engine}, the \emph{Knowledge Base}, the \emph{Reasoner}, the \emph{Interpreter} and the  \emph{Actuation Engine}.



The \emph{Knowledge Base} is responsible for representing aspects of the domain under consideration (such as objects or concepts, instances and states) and their relations, in well defined, machine processable syntax and unambiguous semantics.  The format of the knowledge stored in the knowledge base complies with the knowledge model introduced in Section~\ref{knowledgeModel}, and one of the possible formats is RDF/OWL.  In addition, the knowledge base contains \emph{meta-reasoning expertise}, as well as  \emph{model transformation rules} that are described below.  

The \emph{Reasoner} is used for solving reasoning problems. By reasoning we mean solving problems related to planning, verification, optimization, etc. By relating a \emph{user query} to a \emph{meta-reasoning expertise\footnote{Meta-reasoning is reasoning about reasoning, i.e., it is comprised of computational processes concerned with the operation and regulation of other computational processes within the same entity~\cite{wilson2001encyclopedia}.}} stored in the knowledge base the \emph{Inference Engine} draws a conclusion about an appropriate method or a problem solver, and relevant prior knowledge for solving a given problem. For example, if the user query is to reach a certain goal, the inference engine will look up the knowledge base and deduce that a \emph{Planner} should be used to generate a strategy to reach that goal. Additionally, given that most of the planners accept planning problems in Planning Domain Definition Language (PDDL)~\cite{mcdermott1998pddl} as their input, the inference engine looks up corresponding \emph{model transformation rules} that should be applied to formulate the problem in a format understandable by the selected planner. The \emph{Interpreter} takes the generated strategy and maps it to state changes that it gives to the \emph{Actuation Engine}, so that it can perform actuation in the real world.

Since the physical world is not entirely predictable \kmafShortName{} needs to take into consideration that there might be changes in the information stored in the knowledge base. The \emph{Perception Engine} is responsible, when needed, to push new knowledge from the environment (i.e., predicates) in the knowledge base. Additionally, when executing the strategy the \emph{Interpreter} works tightly with the \emph{Reasoner}. In case there are any changes in the expected state of the system the \emph{Reasoner} sends a replanning request to the \emph{Planner}.

\section{\uppercase {Taxonomy of ITS domain concepts}}
\label{domainModeling}
\noindent This section describes the objects or the concepts of ITS domain and their relation to more generic CPS domain concepts. The concepts are organized in an ontology that has multiple layers of abstraction.  We design our ontology by combining high level concepts and cross-domain relationships borrowed from three areas: CPS; Agent-Based Model (ABM); and Systems-of-Systems (SoS). The proposed ontology consists of an Upper Ontology, which contains the CPS, ABM, and SoS concepts and relations and a general ITS Domain Ontology. The general ITS Domain Ontology can be further referenced from ontologies that instantiate transport-domain specific transitions and states, as described in Section~\ref{implementation}.

The objective of breaking the ontology into multiple levels is twofold. First, this approach allows to capture and isolate different levels of properties, attributes and relationships. Higher layers provide broader definitions and more abstract concepts, while lower layers are less abstract and can support specific domains and applications with concepts and relations which might not be present in the upper levels. Second, ontologies are expected to change, grow and evolve as new domains and techniques are contemplated in them \cite{davies2006semantic}. Leaving the more abstract and general concepts in an upper layer, and the more specific ones in lower layers, reinforces the idea that altering the most general concepts should be avoided, making them less likely to suffer constant modifications that could lead to unnecessary changes throughout the ontology. This is important because ontologies often reuse and extend other ontologies. Updating an ontology without proper care can potentially corrupt the others depending on it and consequently all the systems that use it.

\subsection{Upper Ontology}

Upper ontologies should be designed to describe general concepts that can be used across all domains.
They have a central role in facilitating interoperability among domain specific ontologies, which are built hierarchically
underneath the upper and generic layers, and therefore can be seen as specialization of the more abstract concepts.

\begin{figure}[h]  
\centering
	\includegraphics[width=0.43\textwidth,bb=0 0 516 520]{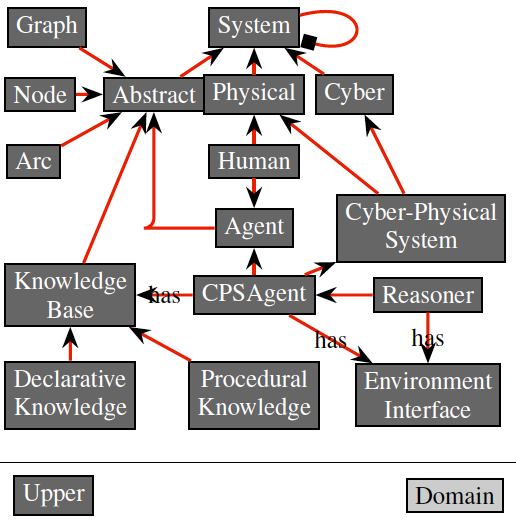}
  \caption{An upper ontology for domain integration.}
 \label{fig:upperOntology}
\end{figure}

Figure~\ref{fig:upperOntology} presents a subset of the proposed upper ontology.
Its development was prompted by our use cases in management and control of complex systems-of-systems,
and was inspired by other ontologies such as SUMO (Suggested Upper Mergerd Ontology)~\cite{niles_towards_2001},
and W3C SSN (Semantic Sensor Network Ontology)~\cite{Compton201225}.

Some important concepts defined on the proposed general ontology include \emph{System}, \emph{Cyber-Physical System}, \emph{Agent}
and \emph{CPS Agent}. A \emph{System} is a set of connected parts forming a complex whole that can also be used as a resource by other systems. A \emph{Cyber-Physical System} is a system with both physical and computational components. They deeply integrate computation, communication and control into physical systems. An \emph{Agent} is a system that can act on its own, sense the environment and produce changes in the world. When an agent is embedded into a cyber-physical system it is called a \emph{CPS Agent}, or cyber-physical agent.

Important for mathematical desciptions of interrelations between systems are the elements \emph{Arc}, \emph{Node} and \emph{Graph}.
Where an \emph{Arc} is any element of a graph that connects two \emph{Nodes}, while a \emph{Graph} is a set of \emph{Nodes} connected by \emph{Arcs}.

The concept of \emph{System} can be further expanded by a number of attributes,
such as \emph{Capacity}, \emph{Role} and \emph{Capability} that can also have relationships
among them. The \emph{System} itself is represented within the \emph{Declarative Knowledge}
as an \emph{Object}. \emph{Affordance} is a property the defines the tasks that can
be done on a specific \emph{System}, while \emph{Capability} defines the set of
tasks the system can perform. Systems can also have \emph{Constraints}, which in turn
are related to \emph{KPI}s that are used to measure whether such constraints are
satisfied.

\begin{figure}
  \centering
  \includegraphics[width=0.43\textwidth,bb=0 0 539 363]{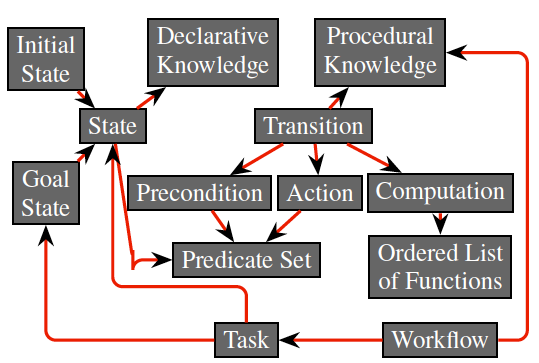}
  \caption{Knowledge Base representation.}
  \label{fig:KnowBasemiddleOntology}
\end{figure}

The higher level of the proposed ontology also provides definitions and relationships between the main \emph{Knowlegde Base}
concepts, the \emph{Declarative} and \emph{Procedural Knowledge}. In our knowledge model, a \emph{Transition}
is a \emph{Procedural Knowledge} concept that determines how to achieve a certain
state (\emph{Action}) given that an agent observes a particular state (\emph{Precondition})
as being true in the world and there is an ordered list of effect free function calls in that state (\emph{Computation}).
Meanwhile, both \emph{Precondition} and \emph{Action} have a \emph{Predicate Set}
that is directly related to the concept of \emph{State} from the \emph{Declarative Knowledge}.
The \emph{Goal State}, which is an specification of \emph{State}, is
related to the concepts of \emph{Task} and \emph{Workflow} from the \emph{Procedural Knowledge}.
Where a \emph{Workflow} is defined as sequence of \emph{Tasks}, which in turn is
defined by a sequence of \emph{Goal States} assigned to a single \emph{Agent}.
Figure~\ref{fig:KnowBasemiddleOntology} presents the main elements of the knowledge
base modeling.

\subsection{ITS Domain Ontology}
\label{ITSDomain}
\begin{figure}[ht!]
	\centering
	  \includegraphics[width=0.43\textwidth,bb=0 0 405 391]{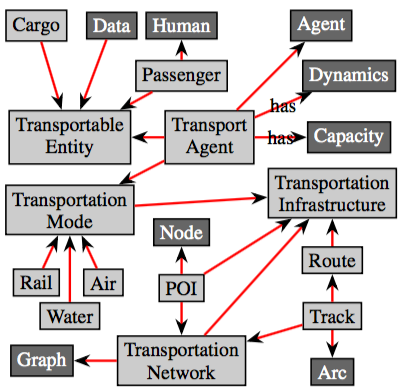}
	\caption{The ITS Domain Ontology.}
	\label{fig:ITSOntology}
\end{figure}

With the support of the presented upper ontology model, in this section we propose an ITS domain
specific ontology, as depicted in Figure~\ref{fig:ITSOntology}.
One of the central concepts within the ITS domain is the
\emph{Transport Agent}, that extends \emph{Agent} from the upper ontology.
The \emph{Transport Agent} encompasses agents that are capable of transporting
some entity, ranging from physical goods to virtual data.
Some important concepts from the upper layers that apply to the \emph{Transport Agent}
include \emph{Dynamics} and \emph{Capacity}, among others. \emph{Transport Agent}s in turn are
strongly related to the \emph{Abstract} concept of \emph{Transportation Mode}
which defines the type of transportion scenario (e.g., \emph{Roads}, \emph{Rail},
\emph{Telco}).

Another important concept is the \emph{Transportation Infrastructure} which encompasses all elements required by a \emph{Transportation Mode}, such as \emph{Route}s, \emph{Track}s and \emph{Transportation Network}s. Most elements within the \emph{Transportation Infrastructure} are extensions of \emph{Graph}, \emph{Arc}
and \emph{Node}, abstract concepts from the Upper ontology. Therefore, by using
high level graph definitions it is possible to define most of the transportation infrastructure
in an ITS Domain. A node inside the transportation infrastructure is referred to as a \emph{POI}
(Point of Interest) and it can be any desired location within the \emph{Transportation Network}
(e.g., a crossing, a specific point in the route, coordinate, a warehouse, a bus stop).
A \emph{Traffic Semaphore} is modeled as a generic \emph{Actuator}
that is used to control and regulate traffic and it can be applied in any
transportation scenario.

A \emph{Transportable Entity} encompasses any element that can be transported
by a \emph{Transport Agent}, such as regular \emph{Cargo} or network \emph{Data}.
A typical \emph{Passenger} is also a \emph{Transportable Entity} and
extends the upper ontology concept of \emph{Human}.


 
\section{\uppercase {Implementation}}
\label{implementation}

\begin{figure*}[!th]
	\centering
	\includegraphics[width=0.7\textwidth]{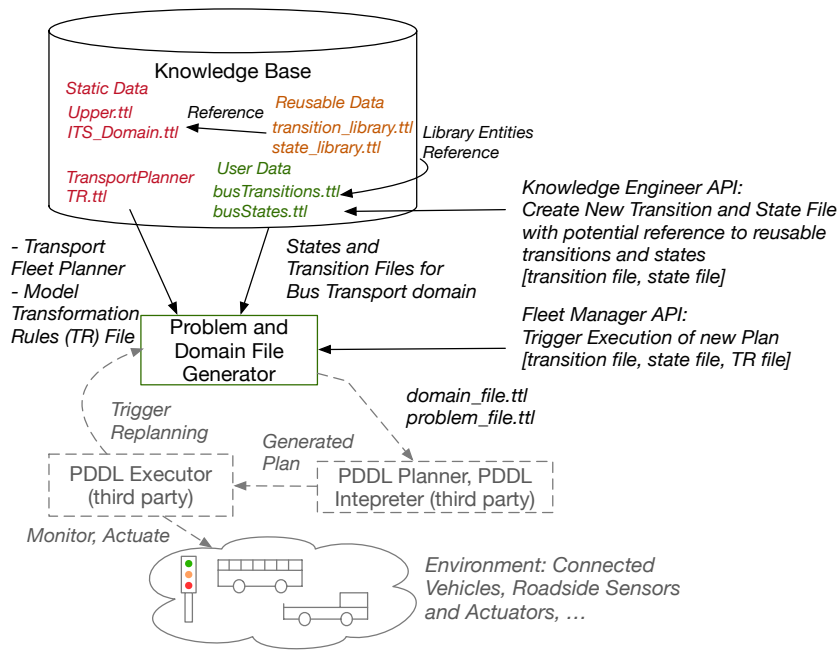}
	\caption{Overview of the implemented system. The ``.ttl" extension denotes model files in Turtle format. Components in dotted rectangles are third-party.}
	\label{implementationFigure}
\end{figure*}

\noindent This section describes current progress towards a prototype implementation of the \kmafShortName{} architecture illustrated in Figure \ref{fig-architecture}. The implementation targets a large problem area in ITS known as ``transportation planning", which we define as \textit{the schedules generated for a set of vehicles to pickup and alight people or cargo along one or more routes, within a given amount of time} (see also Section~\ref{introduction}). The transport planning problem includes a set of connected vehicles, for example buses or trucks, and a central coordinating function that computes the schedule and transmits it to these vehicles\footnote{Correct interpretation of the schedule rests on the vehicles, which can be partially or completely autonomous or they may also have human drivers.}. In this implementation we assume that the \textit{Inference Engine} component has already deduced that a \emph{Planner} should be used to solve the transportation planning problem, using \textit{meta-reasoning expertise} and \textit{user query} data supplied from the \textit{Knowledge Base} and the user interface components respectively. 


Figure~\ref{implementationFigure} shows the components of the implemented system. One of the components implemented is the \textit{Knowledge Base}, which contains \textit{model transformation rules} for PDDL language as well as \textit {states \& transition} models that are based on the structure defined in Section \ref{domainModeling} and contain information for the particular transport planning problem. The models are described using semantic web technologies and are based on the W3C Web Ontology Language (OWL) \cite{pascal_hitzler_owl_2012} and stored in Turtle format \cite{david_beckett_ttl_2014}. The other component is a PDDL Generator, which, given the transformation rules, states \& transition files as input, generates a problem and domain file in PDDL language. PDDL Generator is implemented in Java \cite{gosling_java_2015} and uses Apache Jena \cite{apachejena} for parsing the Turtle-formatted input from the knowledge base. Additionally, Eclipse Jetty \cite{eclipsejetty} provides a Representational State Transfer (REST) API for triggering PDDL file generation, and defining custom states and transition models. More specifically:

\begin{itemize}
\item{The API allows human experts (for example knowledge engineers) to specify a transport logistics problem, by adding a new initial and goal state in the knowledge base, in the form of a \textit{state file}, and a set of transitions with precondition, computation and action parts in the form of a \textit{transition file}. These two files are jointly used by the PDDL Generator software component in order to generate a new schedule. The state file defines the agents, vehicles and routes, contains information about the initial state of the system (e.g., the location of agents in the route, the route and its waypoints, the time required for vehicles to travel a route, etc.) and defines goal conditions (e.g., all agents are serviced). The transitions file describes intermediate transitions that are used by the planner to reach the specified goal state from the initial state. An example of such plan can be found at~\cite{coderepo}.}
\item{The API also provides means for triggering generation of a new PDDL problem and domain file given the above input on request of a human or other system. Typically this request is created from a customer (e.g., human operator, or an automated fleet management system). Once generated, the files are assigned Universal Resource Identifiers (URIs). An external system can subsequently perform Hypertext Transfer Protocol (HTTP) GET requests using the URI references to retrieve the files. An example of such a system can be a PDDL solver\footnote{In its current form, the API does not support adding of new model transformation rules, which means that only PDDL language is supported. In the future however, we plan to expand the functionality by adding support for ``pluggable" problem solving expertise files, both for PDDL and Prolog.}. For this implementation, we use a third-party solver named ``OPTIC", originally developed by Benton et al \cite{ICAPS124699}. }
\end{itemize}

The authors have released the current implementation of the Knowledge Base and PDDL Generator as open-source, available for the community to use~\cite{coderepo}. Currently, there is no component to support interaction with the real world, both for triggering the planning process, but also for actuating real-world connected devices (e.g. buses or sensors) upon execution of the generated plan, however this is planned work. In its current state the implementation can be used for rapid prototyping of transportation planning functions. In addition to the software itself, the ``Upper", ``ITS", ``PDDL model transformation rules" and a set of common reusable transitions and state ontologies are provided in Turtle format. 


\section{\uppercase {Evaluation}}
\label{evaluation}
\noindent In this section, we evaluate the implemented system in terms of reusability. Given the practical limitation of not having a real or simulated testbed of connected vehicles as described in the previous section, it is not yet possible to evaluate some aspects of the system (e.g. efficiency, performance, etc.) in realistic conditions. What we describe instead is an evaluation of the benefits this system brings on ``cost-of-design" (COD). We define COD as the \textit{effort required for formalizing the knowledge required in order for the system to start the automated transport schedule generation process}. Naturally, there exists a direct relationship between the amount of knowledge to be formalized and the effort required, meaning that the more the amount of non-formalized transport plan-related knowledge, the more the required effort (e.g. in terms of time, human resource allocation, money etc.). Table \ref{table:knowledge} shows the different aspects of knowledge required for the transportation plan. As described in Section~\ref{domainModeling}, we view the transport network as a graph, with Points of Interest (POIs) as vertices on the graph and POI-connecting roads as edges. Note that the table references ``transportable entities". These entities can be passengers or cargo, depending on the use case. An interesting observation can be made regarding transitions, which are similar regardless of the route, number and type of transport agents (e.g., buses or trucks) and transportable entities. Therefore, reusing these actions across different transport planning use cases, and storing them as part of the ``transition library" (see Figure \ref{implementationFigure}), is something that can potentially reduce COD.

 \begin{table}[!t]
\renewcommand{\arraystretch}{1.3}
\caption{Knowledge for transport plan generation.}
\label{table:knowledge}
\centering
\begin{tabular}{p{1.8cm}||p{5cm}}
\hline
\bfseries Title & \bfseries Description\\
\hline\hline 
	Route Specification & Specification of the route(s) graph(s), which includes vertices (coordinates), edges (roads) and edge-traversal costs, metrics (e.g. time to travel, fuel spent, etc.)                                                 \\ 
        Transport Agents  & Number of vehicles, Vehicle IDs (e.g. VIN codes) and vehicle capacity                                                                                                                                                  \\ 
        Transportable Entities  & Number and ID of transportable entities                                                                                                                                                                                                \\ 
        Starting Conditions & Where are the vehicles located, where are the passengers located                                                                                                                                                       \\ 
        Transition definitions  & What transitions the transport vehicles perform (which include preconditions, optionally computations and actions) . There are currently three actions available in the library: pickup-agent, drop-agent, move-to-next-coordinate                                                      \\ 
        Goal conditions     & Final state: what criteria needs to be satisfied in order for the transport service to conclude on the specified route (usually this means that all agents are picked up, off-boarded in specific parts of the route). \\
\hline
\end{tabular}
\end{table}

To measure gains in COD, we have defined a simple metric we named ``reusability index". This metric is the ratio of reused entities, versus the ratio of total entities in the knowledge input to PDDL Generator to generate the plan. For the bus use case, this ratio was 0.364 and for the truck use case, the ratio was 0.251. This means that out of the total number entities created for each of the bus and truck use cases, 36.4\% and 25.1\% of them were already available in the library respectively. We observe that both indices are relatively significant, while the difference between the two is mainly attributed to the difference of the route specification entities, as agents in both routes followed different paths. One observation that we have made a posteriori to our measurements, is that the reusability index can be larger, if the route specification is part of the PDDL Generator, which can automatically generate the specification using data from a mapping service in conjunction with a routing library. The knowledge engineer could then only specify the desired waypoints (e.g., the bus stops), and the routes and graphs would be generated automatically by the PDDL Generator. We are currently in the process of evaluating different mapping services such as Google Maps and OpenStreetMap and open-source routing libraries such as GraphHopper and DirectionsService for their applicability in our implementation.
\section{\uppercase {Related Work}}

%
%
%

\noindent A small number of frameworks exist that support the formalization of declarative and procedural knowledge. In~\cite{vaquero_brief_2011} the authors present an overview of tools and methods in KEPS (Knowledge Engineering for Planning and Scheduling) area. They classify the process of knowledge engineering into six phases: Requirements Specification, Knowledge Modeling, Model Analysis, Deploying Model to Planner, Plan Synthesis and Plan Analysis and Post-Design. Then they provide a list of tools and methods that are used in the literature at each of those phases. When commenting on these tools, the authors mention that to that moment none of them treated differently the knowledge encapsulated in the planning problem and in the surrounding domain, which makes it harder to reuse knowledge in other domains. In contrast, the goal of \kmafShortName{} is to exactly support reuse of knowledge, by relying on a common knowledge model across different domains.  


KEWI~\cite{wickler2014ontological} is a knowledge engineering tool that has been designed to help the formalization of a procedural knowledge used for planning problems. The idea is to enable domain experts to encode knowledge themselves, rather than using knowledge engineers. The conceptual model of KEWI consists of three main parts: an ontology for describing entities that occur in the captured domain, a model of primitive rules (called actions) that can be executed by the system and high-level methods for accomplishing complex tasks. KEWI is specifically developed for planning problems, whereas our framework targets different classes of reasoning problems by sharing the same underlying state representation, as formalized in our knowledge model. 


The itSIMPLE tool was first proposed in Vaquero et al.~\cite{vaquero2005itsimple} for supporting the creation of generic planning systems by integrating domain and procedural knowledge, while automatically generating PDDL~\cite{mcdermott1998pddl} files. They propose using UML for modeling domain knowledge and Petri Nets for modeling the procedural knowledge regarding feasible state transitions, while using XML as an internal language. The tool has since evolved~\cite{vaquero_itsimple4_2012} and has been demonstrated in several cross-domain use cases, such as petroleum plant operations planning~\cite{sette_automated_2008}.

In fact, when developing a framework that intends to combine knowledge engineering with multiple declarative and procedural reasoning classes (such as semantic reasoning and planning) one important choice to make is the language (or languages) used to model problems and the required information for solving them.
In~\cite{anis_comparison_2014} the authors compare three classical languages for modeling problems in the cyber-physical production systems, namely: Prolog, Timed Automata~\cite{alur1994theory} and PDDL. Each language has its pros and cons, and can be used for solving different reasoning problems. In \kmafShortName{} we provide a ``glue'' knowledge model that depending on the reasoning problem to be solved is intended to be translated to other languages, such as Prolog, Timed Automata or PDDL. 
Another related field of work is the area of study dealing with cognitive architectures, which are frameworks that specify the underlying infrastructure for an intelligent system and include aspects of a cognitive agent that are constant over time and across different application domains~\cite{langley_cognitive_2009}. As such, they often include memory mechanisms for storing and processing different types of knowledge, e.g., procedural, semantic and episodic knowledge. This knowledge is used as the basis for action selection, which can be reactive, based on planning or a hybrid of these two.
There are many cognitive architectures \cite{samsonovich_toward_2010}, each with its own strategies and constraints when dealing with planning, acting, sensing and knowledge management. Some of the most notorious ones are SOAR~\cite{laird_soar_2012} and CLARION~\cite{licato_structural_2014}. In SOAR, for instance, procedural knowledge is represented by \emph{if-then} rules in a manner somewhat similar to the STRIPS (Stanford Research Institute Problem Solver)/PDDL formalism, while semantic knowledge about the world is represented in a graph held in working memory. CLARION on the other hand employs a hybrid approach by integrating a rule based system with artificial neural networks.
\kmafShortName{} has a number of similarities when compared to cognitive architectures. For instance, \kmafShortName{} intends to improve re-usability across different domains, which is also a natural feature in biologically-based cognitive architectures. Another similarity is in knowledge representation since most cognitive architectures also organize knowledge into either declarative memory or procedural memory. Finally, intelligent agents based on cognitive architectures are also known for performing a range of reasoning tasks such as planning and verification.

In our effort to provide \kmafShortName{} with the capability to reason about problems in different domains, we have included in it a CPS based ontology which describes the relationships among CPS concepts. This approach of using ontologies to increase generality is well known and improves \kmafShortName{}'s capability to reuse knowledge in different domains. One example of related work can be seen in covering Multi-Agent System (MAS) and CPS, as has been explored in~\cite{lin_cps_agents_2010}, where a MAS-based semantic modeling approach to CPS in the water distribution domain was described. The goal was to dynamically integrate information from sensor networks with semantic services to support real-time decision-making. Two challenges were effectively addressed when designing the ontology: model accuracy and integration of physical and cyber components; and pinpointing inter-dependencies between CPS components.

\section{\uppercase {Conclusion}}
\label{section:conclusion}

\noindent In this paper, we have presented concepts and strategies for creating a model for representing knowledge from CPS and ITS domains in a machine-readable form, as well as \kmafShortName{}, a generic framework for knowledge management and automated reasoning, integrating this knowledge model. The motivation behind the creation of the knowledge model and \kmafShortName{} is to demonstrate how knowledge and methods across different domains can be reused to reduce costs of development, deployment and operation of such systems. We have illustrated the usage of parts of \kmafShortName{} on an implementation of a generating transport planner, which can be used to rapidly prototype schedulers of vehicle transport fleets.

There are currently many different abstractions in the literature for handling the design of complex real-world systems, including Cyber-Physical Systems, Agent-Based Model, and Systems-of-Systems engineering, each of them accompanied by their own formalisms and theory. Our vision is that we can provide better support for linking cross-domain use case applications by integrating the common elements of those formalisms via our knowledge model.


In the implementation of \kmafShortName{} we have so far progressed into the development of an ontology for the knowledge base using OWL Web Ontology Language, and we have used the framework to automatically generate PDDL files, feed them into a planner, and create plans. Since, \kmafShortName{} has a much bigger vision than solving planning problems, in the future we plan on studying how our knowledge model can be correlated to other formalisms e.g., Timed Automata.



As future work, we also plan to study how meta-reasoning can help \kmafShortName{} to determine which prior knowledge and algorithms are relevant when a new, problem or unforeseen instance arrives. Such instance corresponds to the current state of the world, along with all current available sensory information. Initially, the system only knows how to solve problems it has seen before and had previously found reductions that could be solved separately using known procedures. If it can not find such a reduction for a new instance, it must recur to space state exploration for generating a sequence of state transitions that either lead to the specified goal state or to a better state in which either the system knows how to proceed with further reductions or declares the problem as intractable under its current knowledge. 






\balance
\bibliographystyle{apalike}
{\small
\bibliography{references}}

\vfill
\end{document}